\documentclass[letterpaper, 10 pt, conference]{ieeeconf}  % Comment this line out if you need a4paper

\IEEEoverridecommandlockouts                              % This command is only needed if 
\usepackage{graphics} % for pdf, bitmapped graphics files
\usepackage{epsfig} % for postscript graphics files
\usepackage{mathptmx} % assumes new font selection scheme installed
\usepackage{times} % assumes new font selection scheme installed
\usepackage{amsmath} % assumes amsmath package installed
\usepackage{amssymb}  % assumes amsmath package installed
\usepackage{bm}
\usepackage{xcolor}
\usepackage{hyperref}
\hypersetup{colorlinks,
            linkcolor=blue,
            citecolor=blue,
            urlcolor=magenta,
            linktocpage,
            plainpages=false}

\title{\LARGE \bf
Sim2Sim Evaluation of a Novel Data-Efficient\\ Differentiable Physics Engine for Tensegrity Robots
}

\author{Kun Wang, Mridul Aanjaneya and Kostas Bekris% <-this % stops a space
\thanks{The authors are with the Department of Computer Science, Rutgers University, NJ, USA. Email:
        {\tt\small {kun.wang2012, mridul.aanjaneya, kostas.bekris}@rutgers.edu}.}%
}

\begin{document}

\maketitle
\thispagestyle{empty}
\pagestyle{empty}

%%%%%%%%%%%%%%%%%%%%%%%%%%%%%%%%%%%%%%%%%%%%%%%%%%%%%%%%%%%%%%%%%%%%%%%%%%%%%%%%
\begin{abstract}
Learning policies in simulation is promising for reducing human effort when training robot controllers. This is especially true for soft robots that are more adaptive and safe but also more difficult to accurately model and control. The sim2real gap is the main barrier to successfully transfer policies from simulation to a real robot. System identification can be applied to reduce this gap but traditional identification methods require a lot of manual tuning. Data-driven alternatives can tune dynamical models directly from data but are often data hungry, which also incorporates human effort in collecting data. This work proposes a data-driven, end-to-end differentiable simulator focused on the exciting but challenging domain of tensegrity robots. To the best of the authors' knowledge, this is the first differentiable physics engine for tensegrity robots that supports cable, contact, and actuation modeling. The aim is to develop a reasonably simplified, data-driven simulation, which can learn approximate dynamics with limited ground truth data. The dynamics must be accurate enough to generate policies that can be transferred back to the ground-truth system. As a first step in this direction, the current work demonstrates sim2sim transfer, where the unknown physical model of MuJoCo acts as a ground truth system. Two different tensegrity robots are used for evaluation and learning of locomotion policies, a 6-bar and a 3-bar tensegrity. The results indicate that only 0.25\% of ground truth data are needed to train a policy that works on the ground truth system when the differentiable engine is used for training against training the policy directly on the ground truth system.

% The simulator introduces a data-efficient linear contact model for accurately predicting collision responses that arise due to contacting surfaces, and a linear actuator model that can drive these robots by expanding and contracting their flexible cables. 

% This engine can be used inside an off-the-shelf, RL-based locomotion controller in order to provide training examples. This paper proposes a progressive training pipeline for the differentiable physics engine that helps avoid local optima during the training phase and reduces data requirements. It demonstrates the data-efficiency benefits of using the differentiable engine for learning locomotion policies for NASA's icosahedron SUPERballBot. In particular, after the engine has been trained with few trajectories to match a ground truth simulated model, then a policy learned on the differentiable engine is shown to be transferable back to the ground-truth model. Training the controller requires orders of magnitude more data than training the differential engine.
\end{abstract}

\section{INTRODUCTION}
Tensegrity robots are composed of rigid elements (rods) and flexible elements (cables), which are connected together to form a lightweight deformable structure. Their adaptive and safe features motivate many applications, such as manipulation~\cite{lessard2016bio}, locomotion~\cite{sabelhaus2018design}, morphing airfoil~\cite{chen2020design} and spacecraft lander design~\cite{bruce2014superball}. While useful and versatile, they are difficult to accurately model and control because of their high number of degrees of freedom and complex dynamics. Identifying system parameters is necessary, either to learn controllers in simulation, as real-world experiments are time-consuming, expensive and perhaps dangerous, or for traditional model-based control. 

The movement of tensegrity robots is made possible by collision responses with objects that are in contact, which is facilitated by normal reaction and friction forces. Different object materials result in normal reaction forces ranging from perfectly elastic to inelastic, and the imperfect contact surface can lead to static or dynamic friction forces as well. Since these forces are generally nonlinear and difficult to model, researchers have proposed various first principles~\cite{hunt1975coefficient, marhefka1999compliant, earles1961impact, polycarpou1995two, brown2017contact} to describe them. These dynamical expressions can have different forms and parameters. In the end, they are only approximations of reality. Thus, finding a reasonably simplified dynamical model that minimizes the amount of parameters needed to be learned is essential for data-efficient system identification.

\begin{figure}
% \vspace{-0.7cm}
    \centering
    \includegraphics[height=0.7\linewidth]{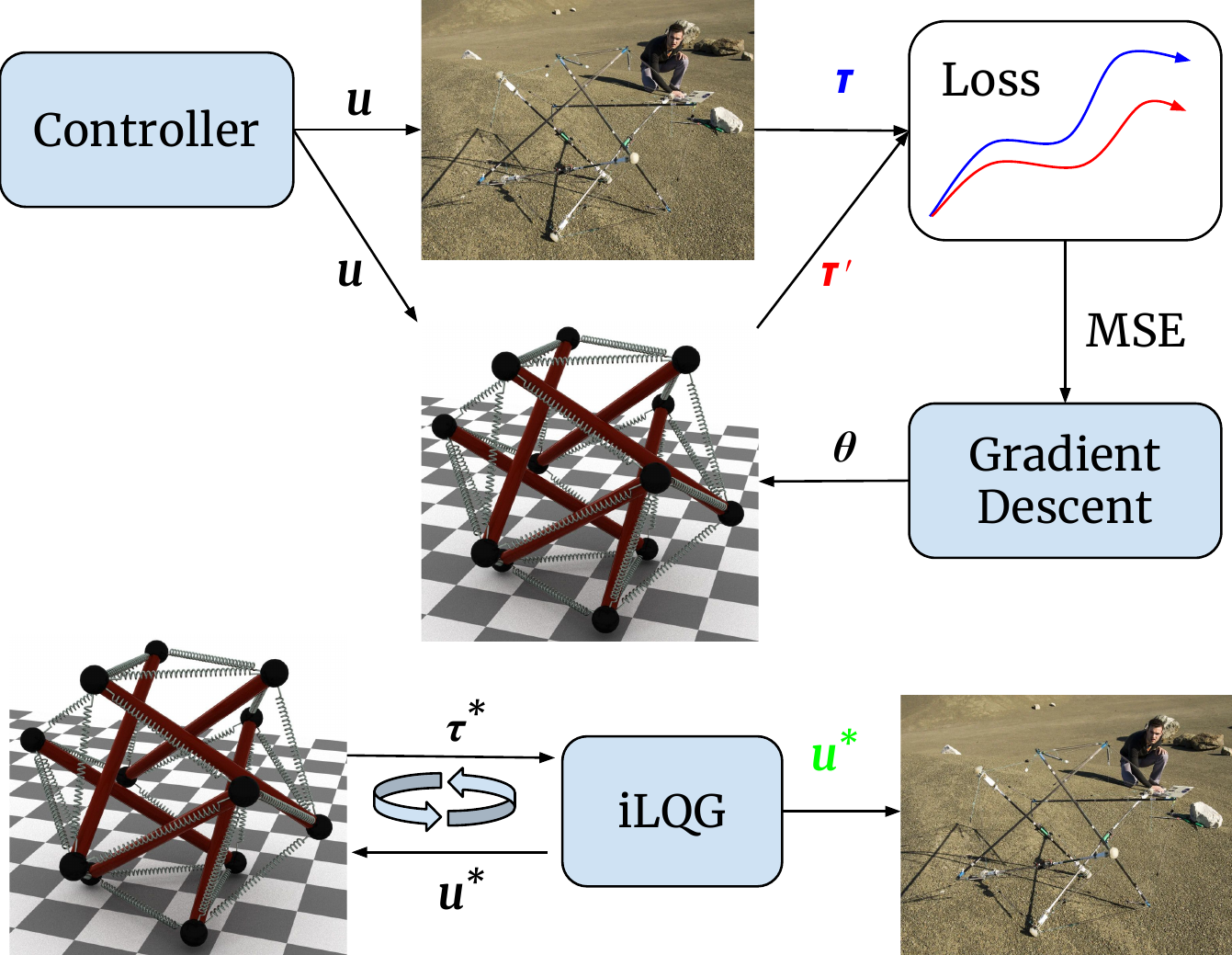}
    % \caption{We proposed a Tensegrity simulator (middle) which could be used for sim2real transfer (left) and sim2sim transfer(right). This work shows successful sim2sim transfer which possible is first and important milestone for sim2real transfer.}
        \caption{The high-level pipeline considered in this work: (top) trajectories from a ground-truth system (here NASA's 6-bar SuperBallBot is shown as an example ~\cite{superball_real_test}) are used to  train a differentiable physics engine so that similar controls results in similar trajectories; (bottom) then a control policy is trained in the differentiable engine and then transferred back to the original system. Primary objectives in this context are minimizing the amount of ground-truth data in the first step and maximizing the effectiveness of the controller in the second step.}
    \label{fig:sim2real_pipeline}
    \vspace{-7mm}
\end{figure}

Differentiable simulation, often based on neural networks, allow for automatic parameter inference from data using backpropagation. They are gaining  attention as a data-driven approach for system identification~\cite{de2018end}. Compared to traditional methods, such as L-BFGS-B~\cite{swevers1997optimal} and CMA-ES~\cite{hansen2001completely}, they often converge faster and are more accurate. Nevertheless, training a differentiable physics engine requires significant amounts of human sampling efforts. This has prompted hybrid designs, where differentiable simulators are combined with analytical models to reduce data requirements~\cite{Wang_FPADSISRS_2020, heiden2020augmenting}.

A differentiable simulator to model cable forces in tensegrity robots was recently proposed ~\cite{Wang_FPADSISRS_2020}. While it has high accuracy and low data requirements, it did not include contact or actuation modeling, preventing  applicability to locomotion. This paper augments this method with a new linear contact model, which can predict complex collision responses. It also includes a new linear actuator model for motors, which applies forces to expand or contract the flexible cables. These models are embedded into the simulation, which leads to an end-to-end differentiable solution for modeling tensegrities and collision responses. A progressive training pipeline is proposed, which helps avoid local optima and reduces data requirements. To the best of the authors' knowledge, this is the \emph{first} differentiable physics engine for such complex robots with all of these features included.  

%Since the ultimate goal is to learn control policies that transfer from simulation to reality with limited ground-truth data, 

This work studies sim2sim transfer as a first step towards sim2real transfer given the complexity of tensegrity robotics. The current sim2sim transfer process shares important challenges of the target sim2real transfer, including very few data from the ground truth system, complex dynamics for the ground truth system that complicate both the system identification and control learning process.

The differentiable engine is evaluated using data sampled from the MuJoCo physics engine~\cite{todorov2012mujoco}, a popular closed-source simulator, which supports rigid body dynamics based on linear complementarity  principles (LCP)~\cite{cottle2009linear} and includes a model for NASA’s icosahedron 6-bar SUPERballBot~\cite{Vespignani:2018:DSCT}. The constraint-based solver inside MuJoCo tries to resolve contact constraints at each time step. MuJoCo does not exactly follow the same first principles contact modeling as that of the proposed differential physics engine. This dissonance makes MuJoCo a good candidate as a ground-truth simulated model to mimic trajectories that would be obtained from a real platform. 

%Besides MuJoCo focusing on general simulation, NTRT simulator~\cite{NTRTSim} was created specifically for tensegrity robots. Although both of these engines provide similar physical realism, tuning their parameters for sim2real transfer of learned controllers requires a lot of human effort \cite{mirletz2015towards, caluwaerts2014design}.

The evaluation section shows that the proposed differentiable physics engine can generate transferable control policies with only $0.25\%$ of the ground-truth data. In particular, after the differentable engine is trained with few trajectories to match the output from MuJoCo, a locomotion policy learned on the differentiable engine is successfully transferred back to MuJoCo.  Training the controller requires orders of magnitude more data than training the differentiable physics engine and is preferable to be executed in simulation.

% The criterion for what constitutes close enough comes from the use of the differentiable simulat as a simulation platform for learning locomotion policies. 

%The main contributions of this work is the introduction of a novel, data-efficient differentiable physics engine for tensegrity robot modeling that significantly reduces the human effort required for collecting ground-truth data and generates transferable locomotion policies. 

Some of the key contributions of the proposed differentiable physics engine for tensegrity robots that allow for data-efficiency and effective system identification are: 
\begin{itemize}
    \item Reasonably simplified contact and actuation models with a few explainable parameters that are amenable to gradient descent optimization.
    \item A progressive identification pipeline that avoids local optima effectively. 
\end{itemize}
Given these insights, the engine is shown effective in sim2sim transfer, where the ground-truth system corresponds to an unknown tensegrity model in MuJoCo, and results in significant reduction in data requirements for successful controller training and transfer.

\begin{figure*}[!htbp]
\vspace{-0.1cm}
    \centering
    \includegraphics[width=0.95\textwidth]{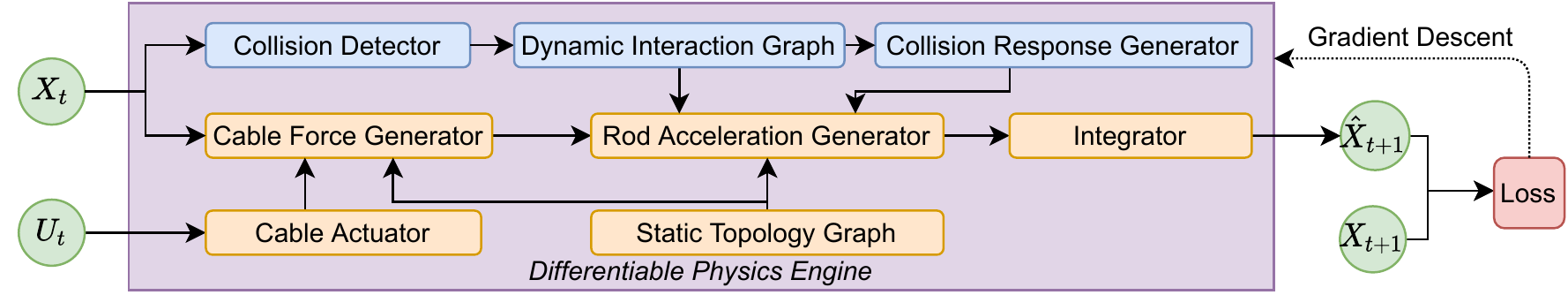}
    \vspace{-3mm}
    \caption{Flow chart showing the training data flow when simulating one time step using our differentiable physics engine.}
    \label{fig:pipeline}
    \vspace{-3mm}
\end{figure*}

\section{RELATED WORK}

The traditional approach for system identification involves building a dynamics model by minimizing the prediction error~\cite{swevers1997optimal,hansen2001completely}. %Alternatively,  \cite{tan2016simulation} used CMA-ES \cite{hansen2001completely} for automatically computing open-loop reference trajectories.
This is achieved by iterating between parameter refinement and data sampling. Data-driven techniques can avoid this iterative process by directly fitting a physical model to data~\cite{rosenblatt1958perceptron,rumelhart1986learning,asenov2019vid2param}. Nevertheless, these techniques are data hungry because they treat the dynamics as a black box, and require retraining in a new environment.

% These methods are general to all systems, whilst \cite{wang2019convex} \cite{fontanelli2017modelling} developed dynamic model identification package specified to the teleoperated surgical robotic system, da Vinci Research Kit(dVRK).

A black box view of the environment can be avoided  by modularizing objects and their interactions in an \emph{interaction network}~\cite{battaglia2016interaction}. A more general hierarchical relation network has been proposed~\cite{mrowca2018flexible} for graph-based object representations of rigid and soft bodies by decomposing them into particles. A Hamiltonian network achieves better conservation of an energy-like quantity without damping~\cite{NIPS2019_9672}. These methods still follow a black-box approach regarding the interactions between different physical elements. This is often a data hungry direction and does not utilize well-understood governing equations.

%newly added!
Differentiable physics engines have received a surge in interest for many applications, including for the prediction of forward dynamics of articulated rigid bodies~\cite{heiden2019interactive}, the solution of inverse dynamics problems for soft materials using the Material Point Method (MPM)~\cite{hu2019chainqueen}, linear complementarity problems (LCP) for multi-body contacts~\cite{de2018end}, and  nonlinear optimization with the augmented Lagrangian method~\cite{landry2019differentiable}. Differentiable engines have also been used with traditional physics simulators for control~\cite{sain}. Differentiable physics engines that are specific to particular objects have been recently investigated, such as molecules~\cite{jaxmd2019}, fluids~\cite{spnets2018}, and cloth~\cite{liang2019differentiable}. 

Prior work by the authors on tensegrity locomotion~\cite{surovik2019adaptive, littlefield2019kinodynamic, surovid2018any} has achieved complex behaviors even on uneven terrain using the NTRT simulator~\cite{NTRTSim}, which has been manually tuned to match a real platform \cite{mirletz2015towards, caluwaerts2014design}. While the locomotion policies have been demonstrated on a real system after trained in NTRT, the sim2real transfer of locomotion policies was quite challenging. This line of work motivated the authors to mitigate the reality gap for cable driven robots. The current work is the first step in this direction by showing sim2sim transfer, where the different and unknown physical model of MuJoCo acts as the ground truth.

\section{PROBLEM FORMULATION}

The objective is to design a predictive dynamics model for tensegrity robots using the governing equations of motion as first principles. The model must be able to return the next state $X_{t+1}$ using knowledge of the current state $X_t$ and control signals $U_t$. The focus is on modeling the entire state space of cable-driven tensegrity robots, which includes rigid elements, flexible cables, complex contacts with the environment, and associated non-linearities and oscillations. 

Since analytical solutions that express the dynamics of such systems either employ approximations that do not hold in the real world, or require knowledge of parameters (such as friction, restitution coefficient, mass, etc.) that are unknown, this work adopts a (model-based) machine learning perspective. The available data are pairs of the form $[(X_t, U_t), X_{t+1}]$, which are collected by observing and decomposing trajectories sampled from a target, ground-truth platform, which in this paper come from MuJoCo~\cite{todorov2012mujoco}. The current work focuses on addressing the high-dimensionality and complex dynamics of tensegrity robots. It assumes that the robot's state is directly observable as a stepping stone towards addressing the general case where sensing data are used to indirectly and imperfectly observe the system's state. 

The metric of success for the above objective is whether the differentiable physics engine allows the learning of locomotion policies using an off-the-shelf controller by using fewer ground-truth trajectories than would be otherwise required (i.e., \emph{without} the differentiable physics engine). In particular, consider the following scenario:
\begin{itemize}
\item An RL-based controller requires $C_{RL}$ amount of ground-truth data of the form $[(X_t, U_t), X_{t+1}]$ to solve a task with the target platform, with a performance score $P_{RL}$. 
\item The amount of ground-truth data required to train the differentiable physics engine is $C_{PE}$. 
\item The controller can also be used to learn locomotion policies using the differentiable physics engine and fewer $C_{GT}$ ground-truth data. It achieves a performance score $P_{PE}$ on the target platform in this case.  
\end{itemize}
Then, the metric of success for this work is that $C_{PE} + C_{GT} < C_{RL}$ and $P_{PE} \geq P_{RL}$. This means that the controller requires fewer ground-truth data to be trained in conjunction with the differentiable physics engine, in contrast to when it is trained directly on the ground-truth system. Moreover, it should achieve at least the same level of performance. 

The experiments accompanying this paper use the SUPERballBot platform~\cite{Vespignani:2018:DSCT}, as simulated in the MuJoCo engine as the target, ground-truth platform. Its state $X_t$ is 72-dimensional since there are 6 rigid rods for which the 3D position $\boldsymbol{p_t}$, linear velocity $\boldsymbol{v_t}$, quaternion $\boldsymbol{q_t}$ and angular velocity $\boldsymbol{\omega_t}$ are tracked over time. The space of control signals $U_t$ is 24-dimensional, since there are 24 cables and it is possible to set the target length for each of them individually. For the learned controller, this work employs iterative LQG (iLQG)~\cite{todorov2005generalized}, used before to learn locomotion policies for this tensegrity~\cite{surovid2018any,surovik2019adaptive}, where the objective was to achieve a desired speed for the platform's center of mass.

\section{METHODS}

\subsection{Differentiable Physics Engine}

Figure ~\ref{fig:pipeline} provides the architecture of the proposed differentiable physics engine. The loss function is the L2-norm of the difference between the predicted next state $\hat{X}_{t+1}$ and the ground-truth state $X_{t+1}$ at time $t+1$. The internal parameters of the physics engine are optimized using backpropagation with gradient descent. The engine includes multiple modules, which fall into two categories: a) the contact simulation modules, and b) the tensegrity robot simulation modules. Note that this model is an approximation of the real dynamics. For example, the rods and springs are coupled together but the differentiable physics engine models them independently. This choice is reasonable because shocks originating from contacts with rigid rods are absorbed by the compliant springs. Moreover, as Section~\ref{sec:experiments} shows, this setup converts the otherwise highly non-linear dynamical system into a simpler system whose parameters can be efficiently learned through a data-driven approach.

\begin{figure}[h]
\vspace{-3mm}
    \centering
    \includegraphics[width=0.8\linewidth]{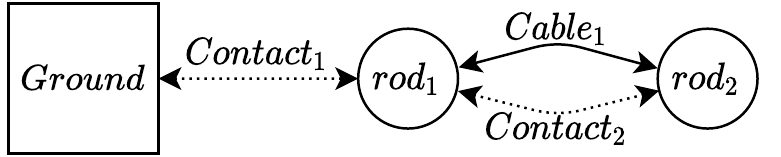}
    % \vspace{-8mm}
    \caption{Relation graph: dynamic interaction graph and static topology graph for a system with two rods (which may collide), one cable, and the ground.}
    \label{fig:relation_graph}
    \vspace{-2mm}
\end{figure}

The contact simulation modules include: 1) a collision detector, 2) a dynamic interaction graph, and 3) a collision response generator. The collision detector rebuilds the world-space configuration of the system by using the current state $X_t$ and an off-the-shelf collision detection algorithm, which outputs contact information, including the colliding object pairs, collision point and intersection depth. This information is used for computing a \emph{dynamic interaction graph}, which describes the contact relations between the different objects. Figure~\ref{fig:relation_graph} illustrates a simple environment with two rods, one cable and the ground. The bidirectional connectors reflect relations between objects. In particular, the dashed connectors denote the contact relations, which are dynamic during the simulation. The collision response generator predicts collision forces and torques for each colliding pair of objects.

Following prior work~\cite{Wang_FPADSISRS_2020}, the robot simulation modules include 1) a static topology graph, 2) a cable actuator, 3) a cable force generator, 4) a rod acceleration generator, and 5) an integrator. The static topology graph shows how cables and rods are connected together, and where the connection anchors are. The solid connections in Figure~\ref{fig:relation_graph} denote the static topology of the robot's components. The cable actuator controls the cable's \emph{rest length} with the control signal $U_t$. The cable force generator is an analytic model based on Hooke's law. The rod acceleration generator takes forces and torques as inputs and maps them to the rod's linear and angular accelerations using the relation graph. Finally, the integrator applies the semi-implicit Euler method to compute $\hat{X}_{t+1}$.

\subsection{Collision Response Generator}
The collision response generator produces contact forces based on contact states computed using the dynamic interaction graph.  Each contact state includes the contact point, contact depth $d$, contact normal $\boldsymbol{n}$, and the indices of the two contacting objects, which have relative linear velocity $\boldsymbol{v}$ at the contact point. With $d$, $\boldsymbol{n}$ and $\boldsymbol{v}$, the module computes the contact forces including the reaction force, sliding friction force, torsional friction force, and rolling friction force.

\subsubsection{Reaction Force}
The reaction force $\boldsymbol{f_r} = j\boldsymbol{n}$ is derived using an impulse-based contact model~\cite{Goldstein:2001:CM}. The reaction force formula is: \vspace{-.1in}
\begin{align}
        j = \dfrac{ -\beta d /\Delta t^2 -( 1 + e ) \boldsymbol{v} \cdot \boldsymbol{n} / \Delta t } { m_1^{-1} + m_2^{-1} + (( \boldsymbol{I_1^{-1}} ( \boldsymbol{r_1} \times \boldsymbol{n} ) \times \boldsymbol{r_1} )  +  ( \boldsymbol{I_2^{-1}} ( \boldsymbol{r_2} \times \boldsymbol{n} ) \times \boldsymbol{r_2} )) \cdot \boldsymbol{n}} \vspace{-.1in}
\end{align}
where $\beta$ is the bias factor, $e$ is restitution coefficient, $\Delta t$ is the simulation time step, $m_1, m_2$ are the masses of the objects, $\boldsymbol{I_1, I_2}$ are their inertia tensors in world frame, and $\boldsymbol{r_1, r_2}$ are the torque arms of the two objects at the collision point.

Inspired from Baumgarte stabilization~\cite{Featherstone:1987:RDA}, the bias factor $\beta\in[0,1]$ accounts for the percentage of error, i.e. $d$, that is corrected at each time step with the collision response.

The coefficient of restitution $e$ relates the pre-collision relative velocity $\boldsymbol{v}$ at the contact point to the post-collision relative velocity $ \boldsymbol{v'}$ along the contact normal $\boldsymbol{n}$ using the relation $\boldsymbol {v'}\cdot \boldsymbol {n} =-e\boldsymbol {v} \cdot \boldsymbol {n}$. Prior methods have used $e\in [0, 1)$ for modeling hard contacts, showing that $\boldsymbol{v'}$ is in the opposite direction of, and less than, $\boldsymbol{v}$. Nevertheless, the choice made here is to set $e\in (-1, 1)$, accounting for the case when $\boldsymbol{v'}$ is only less than, but in the same direction as, $\boldsymbol{v}$. This allows to model adhesive contacts if $1+e \in (0, 1)$.

The only parameters that require system identification are $\beta$ and $e$. Since both of them are within the interval $[-1, 1]$, they can be conveniently identified during the training process using the same learning rate that is used by all other parameters for updating the gradients. Although the additional terms $1/\Delta t^2$, $1/\Delta t$ do arise from physical first principles, they also benefit the optimization during training. Since the magnitudes of $d$ and $\boldsymbol{v}$ are different, $1/\Delta t^2, 1/\Delta t$ help to normalize them to the same interval.

The linear model can also be viewed as an improved version of the Kelvin-Voigt linear model~\cite{Goldstein:2001:CM}:
\begin{align}
    \boldsymbol{f_r} = (-ad-b\boldsymbol{v}\cdot \boldsymbol{n})\boldsymbol{n}
\end{align}
where $a$ is the stiffness, $b$ is the damping coefficient. A concern with the Kelvin-Voigt model is that the transition between contact and non-contact conditions is discontinuous due to the use of a linear dashpot. The damping parameter may also result in a non-physical negative normal force during separation~\cite{brown2017contact}. Parameters $a$ and $b$ also have different magnitudes, and a unique learning rate may find it difficult to update their gradients. This model doesn't consider the effects of collision torques either. Our linear model, however, changes this situation, since all parameters can use the same learning rate, and torques are considered as well. 

\subsubsection{Sliding Friction Force}
Sliding friction opposes relative sliding between the two objects. The sliding friction is derived as $\boldsymbol{f_s} = k \boldsymbol{t}$ from the impulse-based contact model~\cite{Goldstein:2001:CM}:
\begin{align}
        k = \dfrac{ -(1 - \gamma) \boldsymbol{v} \cdot \boldsymbol{t} / \Delta t } { m_1^{-1} + m_2^{-1} + (( \boldsymbol{I_1^{-1}} ( \boldsymbol{r_1} \times \boldsymbol{t} ) \times \boldsymbol{r_1} )  +  ( \boldsymbol{I_2^{-1}} ( \boldsymbol{r_2} \times \boldsymbol{t} ) \times \boldsymbol{r_2} )) \cdot \boldsymbol{t}}
\end{align}
where $\boldsymbol{t}$ is the sliding friction direction, which is opposite to the tangential projection of the relative velocity $\boldsymbol{v}$ onto the contact plane, and $\gamma$ is the damping coefficient.

The coefficient of damping $\gamma$ relates the pre-collision relative velocity $\boldsymbol{v}$ of the contact point to the post-collision relative velocity $ \boldsymbol{v'}$ along the tangential sliding friction direction $\boldsymbol{t}$  as follows $\boldsymbol {v'}\cdot \boldsymbol {t} =\gamma \boldsymbol {v} \cdot \boldsymbol {t}$. The parameter $\gamma$ is set to be in $[0, 1]$. Thus, $\boldsymbol {v'}\cdot \boldsymbol {t}$ can only be in the same direction, but less than $\boldsymbol {v} \cdot \boldsymbol {t}$. 

The sliding friction is governed by Coulomb friction, i.e.,
\begin{align}
   \boldsymbol{f_s} = \begin{cases}
       k \boldsymbol{t} & \text{if $k < \mu ||\boldsymbol{f_r}||$} \\
       \mu ||\boldsymbol{f_r}|| \boldsymbol{t} & \text{if $k \geq \mu ||\boldsymbol{f_r}||$}
   \end{cases}
\end{align}
where $\mu\in [0,1]$ is the coefficient of friction and $||\boldsymbol{f_r}||$ is the norm of the reaction force. The only parameters that require system identification are $\gamma$ and $\mu$. 

\subsubsection{Torsional and Rolling Friction}
The torsional and rolling friction forces oppose the spinning and rolling of the object. Instead of LCP,  a simplified method is used. Since the only object that might spin or roll is the rod (the ground has infinite inertia), and the rod can only roll along its principle $\boldsymbol{z}$ axis, a frictional torque is applied along its $\boldsymbol{z}$ axis to mimic the effects of these two friction forces.
\begin{align}
    \tau_f = (\epsilon - 1) \boldsymbol{I}^{-1} ((\boldsymbol{\omega} \cdot \boldsymbol{z})\boldsymbol{z})/\Delta t,
\end{align}
where $\boldsymbol{\omega}$ is the object's angular velocity, and $\epsilon$ is the damping coefficient, which relates the pre-collision object angular velocity $\boldsymbol{\omega}$ to the post-collision angular velocity $ \boldsymbol{\omega'}$ as follows $\boldsymbol {\omega'}\cdot \boldsymbol {z} =\epsilon \boldsymbol {\omega} \cdot \boldsymbol {z}$. By setting $\epsilon\in [0, 0.01]$ to only allow minor twist, then $\boldsymbol {\omega'}\cdot \boldsymbol {z}$ could only be in same direction, but less than $\boldsymbol {\omega} \cdot \boldsymbol {z}$. The only parameter that requires system identification is $\epsilon$. 

In summary, the collision response generator only has three linear models with 5 parameters to identify, $\beta, e, \gamma, \mu, \epsilon$.

% The contact response generator is a deep contact neural network (DCNN), shown in Figure~\ref{fig:collision_response_generator}, which is a multi-layer perceptron (MLP). The inputs are contact states $X_{c}$, including intersection depth $d$, torque arms $r_c$, contact point $p_c$, colliding objects states $S_i, S_j$ ($S$ includes position, quaternion, linear and angular velocity). The outputs are contact forces $f_c$, including reaction and tangential friction forces and contact torques $\tau_c$, including torsional and rolling frictions. We didn't apply first principles here since the responses are highly nonlinear, thus these principles can't converge to reasonable formulas, especially the linear ones.

% \begin{figure}
% % \vspace{-0.7cm}
%     \centering
%     \hspace*{-5mm}
%     \includegraphics[width=1.2\linewidth]{figures/collision_response_generator.pdf}
%     % \vspace{-8mm}
%     \caption{Data flow of collision response generator}
%     \label{fig:collision_response_generator}
%     % \vspace{-5mm}
% \end{figure}

\subsection{Cable Actuator}
The tensegrity robot has actuators at the end of each cable, which are motors with a reel that cast and retract the cables. The control signal $u$ is the target position that the actuator must roll to. The actuation forces are computed as follows:
\begin{align}
    w_{t} &= w_{t-1} + a (u_{t-1} - w_{t-1}) \\
    \Delta l_{t}^{rest} &= b w_{t} \\
    l_{t}^{rest} &= l^{rest} + \Delta l_{t}^{rest} 
\end{align}
where $a$ is the motor speed, $w$ is the motor position, $b$ is a scalar mapping the motor position to the change in the cable's rest length, and $l^{rest}$ is the cable's rest length when the motor position $w=0$. The only parameters that require system identification are $a$ and $b$. The cable's rest length $l^{rest}$ is updated by the motor position $w$ at each time step.

\subsection{Progressive Training}
This work applies progressive training to avoid local optima. Combining tensegrity and contact simulation, the system dynamics are highly nonlinear and have many parameters to identify. This makes the training loss landscape highly nonconvex and gradient descent often converges to parameter estimates that poorly describe the ground-truth data~\cite{heiden2020augmenting}. Parallel random search could be a possible solution but is computationally expensive. Instead of identifying the whole system in one shot, the method \emph{divides} all system parameters into groups and identifies these groups \emph{progressively}. It also prioritizes the identification for groups of parameters that have higher fidelity. 

The first step is to identify non-contact parameters, such as the actuator speed, actuator position scalar, cable stiffness, cable damping, rod mass, etc., which only describe the robot itself. The tensegrity is first help up in the air, actuated with random controls and non-contact trajectories are collected to identify non-contact parameters.

Then these non-contact parameters are fixed and the next step is to identify contact parameters in a similar, progressive manner: (i) first train the linear contact model and then  \emph{fix} it; (b) then, augment the linear model with a Multi-layer Perceptron (MLP) to learn complex dynamics.

\subsection{Implementation}
The engine is implemented using Pytorch, which provides built-in support for auto-differentiation. The range of parameters are limited by activation functions, e.g., the sigmoid and  hyperbolic tangent functions. The learning module receives the current state $X_t$ and returns a prediction $\hat{X}_{t+1}$.  The loss function is the mean square error (MSE) between the predicted $\hat{X}_{t+1}$ and the ground-truth state $X_{t+1}$. The proposed decomposition, first-principles approach, and the cable’s linear nature allow the application of linear regression, which helps with data efficiency. This linear regression step has been implemented as a single layer neural network without activation functions.

%%%%%%%%%%%%%%%%%%%%%%%%%%%%%%%%%%%%%%%%%%%%%%%%%%%%%%%%%%%%%%%%%%%%%%%%%%%%%%%%
\section{EXPERIMENTS}
\label{sec:experiments}

The setup of MuJoCo is listed as follows. The rod mass is 10kg. The rod's length is 1.684m and cable rest length is 0.95m. The cable stiffness is 10000 and the cable damping is 1000. The cable actuator has control range $[-100, 100]$. The activation dynamics type is filter. The gain parameter is 1000. The dimensionality of the contact space is 6, including sliding, torsional and rolling friction. The torsional and rolling coefficient is set to 1. The integrator is set to the semi-implicit Euler method.

\subsection{Tensegrity Robot Model Identification}

\begin{figure}[h]
% \vspace{-0.7cm}
    \centering
    \includegraphics[height=0.37\linewidth]{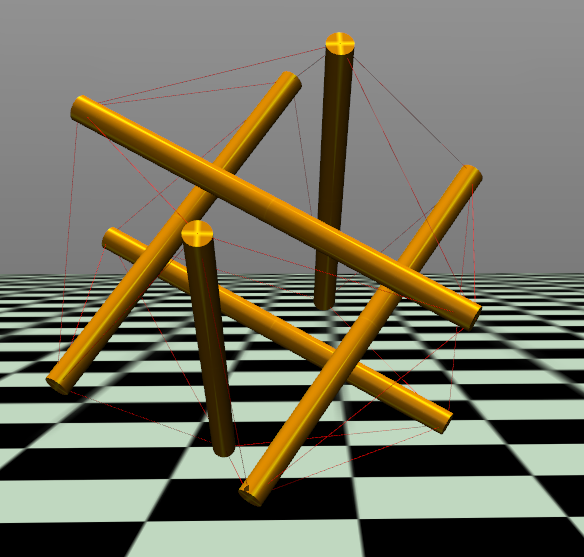}
    \includegraphics[height=0.37\linewidth]{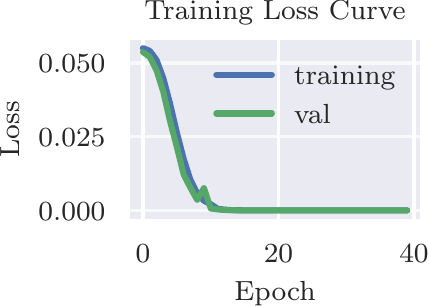}
    \caption{(left) The robot is placed in the air and random controls are executed to sample non-contact trajectories. (right) Training loss curve after the model identification process.}
    \label{fig:superball_non_contact_identification}
    \vspace{-7mm}
\end{figure}

As explained, the first step is to identify parameters of the tensegrity robot itself without considering contacts, i.e., rod mass $m$, cable stiffness $k$ and damping $c$, actuator motor speed $a$ and position scalar $b$. One rod of the SUPERball is fixed in the air and all 24 cables are actuated with random controls, as shown in Figure~\ref{fig:superball_non_contact_identification} (left). One trajectory is collected for training, one for validation and 10 for testing. Each trajectory lasts for 5 seconds, i.e., 5,000 time steps as the simulation step $\Delta t$ is 1ms. 

The training uses an Adam optimizer for 40 epochs, 256 batch size. The initial learning rate is 0.1, and is reduced by half every 10 epochs. Both the training loss and validation loss go down from 0.055 to 3e-11. Figure~\ref{fig:superball_non_contact_identification} (right) shows the stable training curves.

From Newton and Hooke's law, it is possible to infer that: $a = -(k/m)\Delta x - (c/m) \Delta v.$ Thus, any parameter combination satisfying $K/m = 10000/10=1000$ and $c/m = 1000/10 = 100$ is a solution. Instead of evaluating the absolute parameters $K, c$, the result report the relative ratios $K/M$ and $c/M$. The identified parameters are $m=0.03191, k=31.917, c=3.1917, a=0.001, b=0.1$, thus $k/m = 31.917/0.03191=1000, c/m=3.1917/0.03191=100$, which are identical to the ground truth. Combining semi-implicit integration, these parameters produce the exact same motions for the robots.

\begin{figure}[h]
\vspace{-0.1in}
    \centering
    \includegraphics[width=0.49\linewidth]{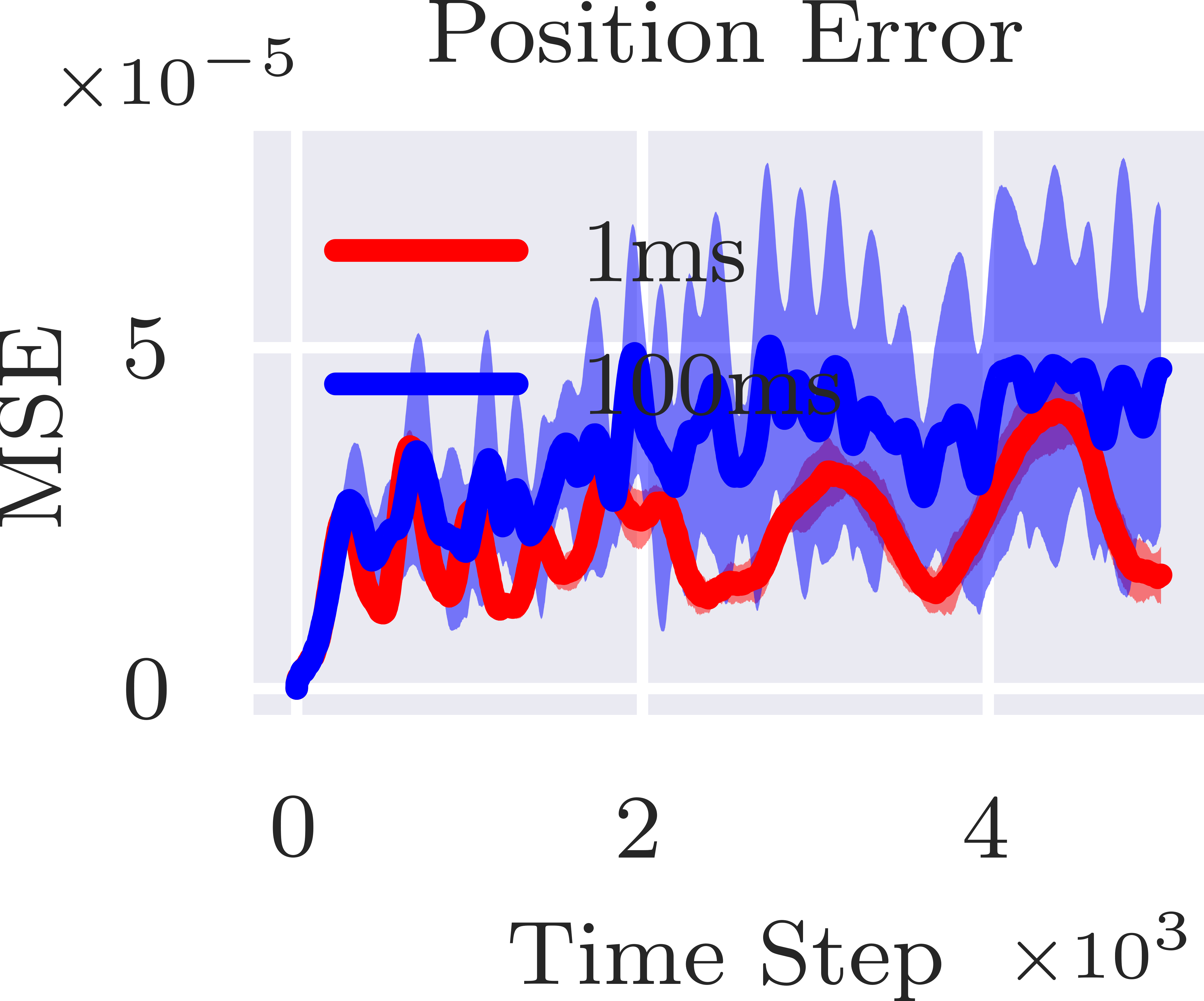}
    \includegraphics[width=0.49\linewidth]{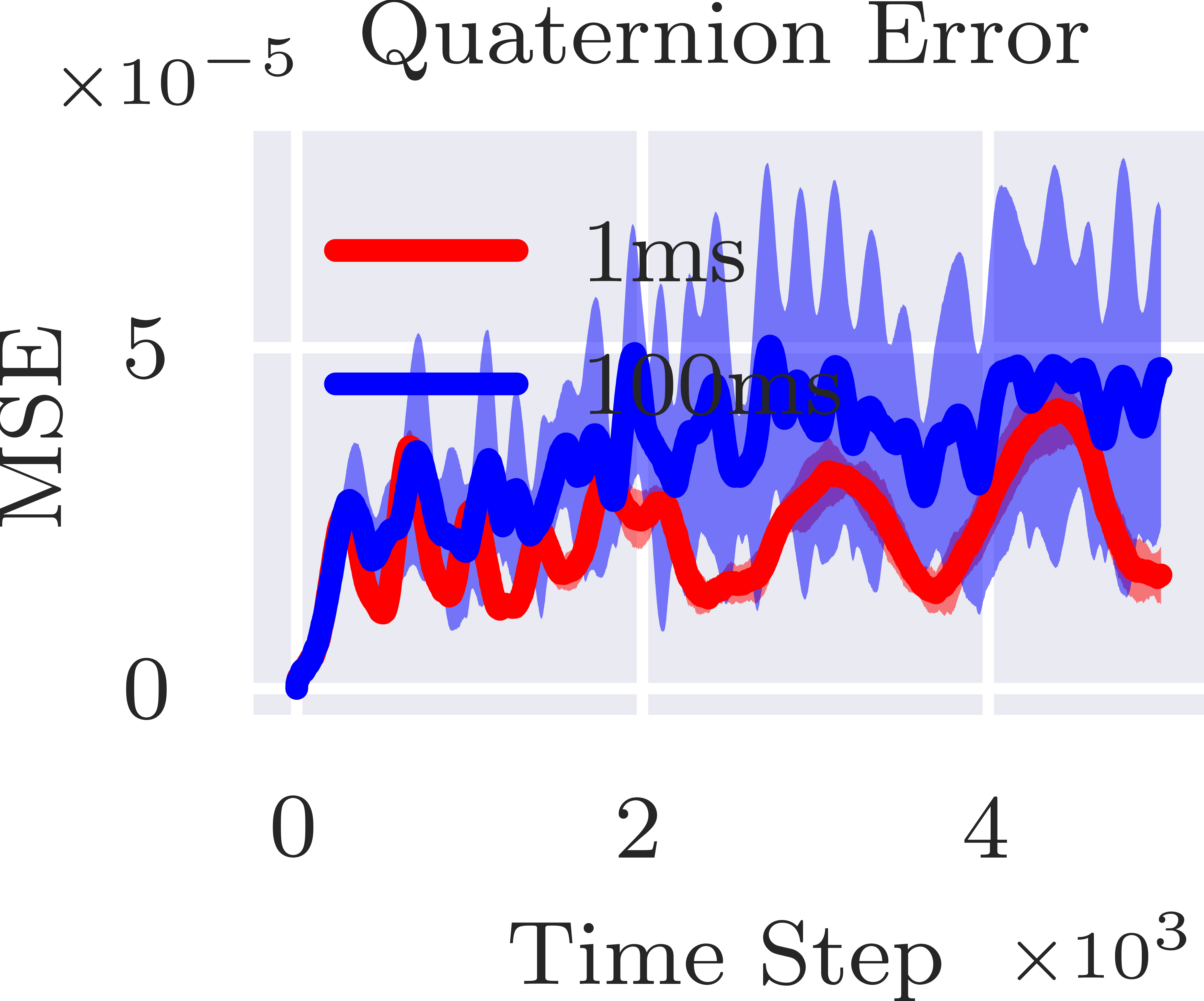}
    \vspace{-8mm}
    \caption{Without contacts, the identified robot model is very accurate in terms both of (left) position  and (right) orientation error. The thick line means the average and transparent region means the variance.}
    \label{fig:superball_non_contact_identification_error_graph}
    \vspace{-1mm}
\end{figure}

The identified parameters are tested on two datasets. One includes 10 trajectories with controls every 1ms and 10 trajectories with controls every 100ms. The controls are randomly sampled from the uniform distribution in range $[-100, 100]$. 100ms controls are used because 100ms is a popular control interval for real world setups.  Figure~\ref{fig:superball_non_contact_identification_error_graph} shows that the model performs good on both of them. After 5,000 time steps, it achieves both an error in the order of $10^{-5}m$. 

\subsection{Tensegrity Contact Model Identification}
\begin{figure}[h]
\vspace{-0.1in}
    \centering
    \includegraphics[width=0.49\linewidth]{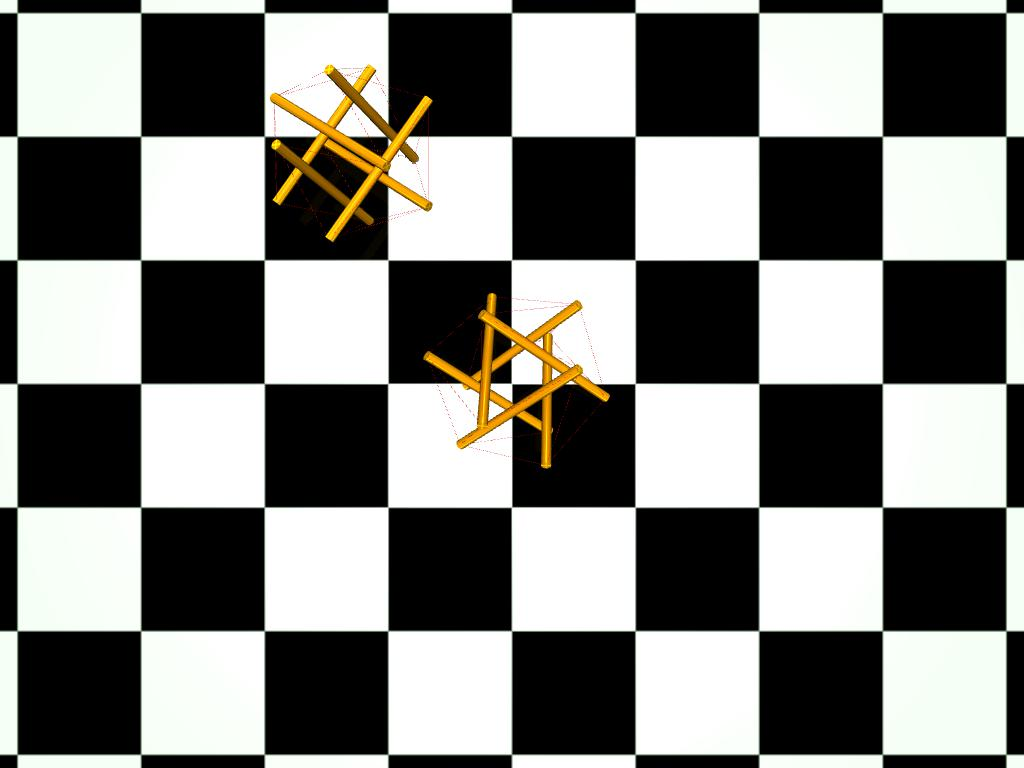}
    \includegraphics[width=0.49\linewidth]{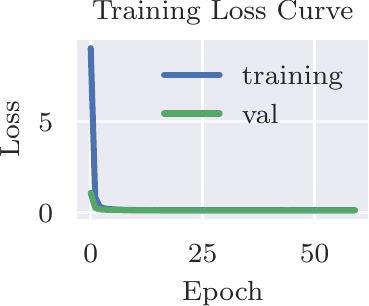}
    \vspace{-5mm}
    \caption{The rolling trajectories are sampled with random initial directions and  speeds (left). These trajectories are trained by our contact models with gradient descent (right).}
    \label{fig:linear_model_1_traj_training}
    \vspace{-2mm}
\end{figure}

Given the identified non-contact parameters, the contact parameters to estimate are $\beta, e, \gamma, \mu, \epsilon$ for the collision response generator. The SUPERball is thrown to a random direction with a random initial speed. The SUPERball will start rolling on the ground and finally stop. One such trajectory is used for training, 1 for validating and 10 for testing. Each trajectory is 5 seconds long, i.e. 5,000 time steps. The environment setup and the training loss curve are shown in Figure~\ref{fig:linear_model_1_traj_training}.

The training uses an Adam optimizer for 60 epochs, 256 batch size. The initial learning rate is 0.1, and is reduced by half every 20 epochs. Both the training loss and validation loss go down to 0.13. Compared to the non-contact training process, the minimum training error here, 0.13, is larger than the maximum training error before, 0.055. This shows the necessity of the progressive identification pipeline. Figure~\ref{fig:linear_model_1_traj_training} shows that the simplified linear contact model converges fast and stable. It only takes 1 trajectory for training, which is favorable for a real-world setup.

\begin{figure}[h]
\vspace{-0.1in}
    \centering
    \includegraphics[height=0.35\linewidth]{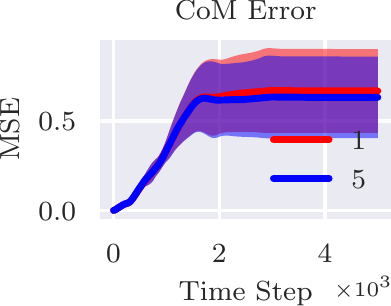}
    \includegraphics[height=0.35\linewidth]{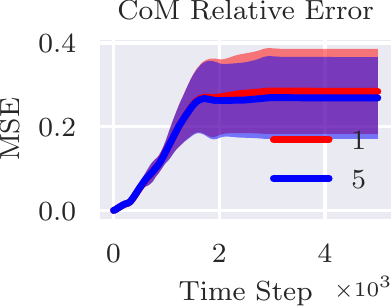}
    \vspace{-2mm}
    \caption{The models trained from 1 and 5 trajectories shows similar performance on the CoM error. More data doesn't result a better model. This reflects the trade-off between model simplicity and prediction accuracy.}
    \label{fig:linear_model_1_vs_5_traj_training}
    \vspace{-2mm}
\end{figure}

The identified parameters are tested on a dataset with 10 rolling trajectories, which start with a random direction and a random initial speed. The evaluation uses the trajectory difference for the Center of Mass (CoM). Figure~\ref{fig:linear_model_1_vs_5_traj_training} shows that the model has error approx. 0.63m after 4,000 time steps. Considering the average trajectory length is 2.35m, the relative error is in the order of 26.8\% its length. The robot starts with a random initial speed and stops in a range from 100 to 2,000 time steps. After 2,000 time steps, the CoM error curve becomes flat because the ball stops rolling. The error arises from the simplification of the contact model. The simplification reduces the training data significantly, but also sacrifices identification accuracy, which is a trade-off between data requirements and model complexity. The objective is not to get the most accurate but also data hungry model. Instead, it is to find a data-efficient, simple model that is sufficient for learning a good policy that transfers back to the ground truth system. Section V.C evaluates the model and shows that the resulting model can still generate effective policies for the ground-truth system. 

Figure~\ref{fig:linear_model_1_vs_5_traj_training} shows that testing error doesn't decrease when the training dataset includes 5 trajectories. This is because the contact model only has 5 parameters, however the ground truth model is more complex. This motivates the use of the neural network to learn  more complex dynamics.

\subsubsection{Evaluation of an MLP}
An MLP (Multi-layer Perceptron) is a popular neural network model to learn non-linear functions. The first test is to fully replace the proposed model with an MLP, which receives contact states and predicts contact forces and torques. The input of the MLP is a 29 dimension vector, based on contact principles, including $t_c, d, \boldsymbol{v}, \boldsymbol{r_1}, \boldsymbol{r_2}, \boldsymbol{r_1 \times n}, \boldsymbol{r_2 \times n}, \boldsymbol{z_1}, \boldsymbol{z_2}, \boldsymbol{\omega_1}, \boldsymbol{\omega_2}$ ($t_c$ is the time step during collision). To use less training data, all input elements are transformed into the contact frame. MLP has 3 fully connected layers and each layer has 100 hidden variables.%The contact frame is shown in Figure~\ref{fig:contact_local_frame}. 

\begin{figure}[h]
\vspace{-0.1in}
    \centering
    \includegraphics[width=0.49\linewidth]{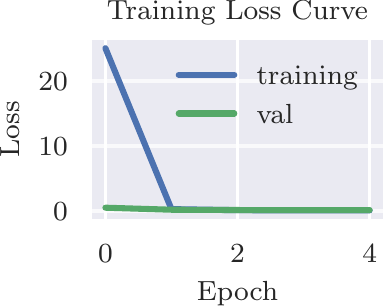}
    \includegraphics[width=0.49\linewidth]{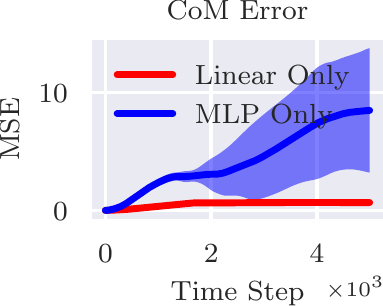}
    \vspace{-6mm}
    \caption{Training a MLP to replace our linear contact model results worse performance. Although the training loss is low, the large testing errors indicate the overfitting.}
    \label{fig:superball_mlp_contact_model}
    \vspace{-2mm}
\end{figure}

The MLP is trained for 5 epochs, 256 batch size. The initial learning rate is 0.1, and is reduced by half every 1 epoch. Fewer epochs are used and the learning rate is reduced quickly since the MLP can overfit easily to a small training dataset. The training and validation loss is shown in Figure~\ref{fig:superball_mlp_contact_model} left. The performance comparison is in Figure~\ref{fig:superball_mlp_contact_model} right. Although the validation loss keeps steady during the training process, the testing error indicates that MLP overfits the training data. In a small dataset, both training and validation dataset are biased and the validation loss is not that helpful to indicate whether overfitting takes place.

\subsubsection{Augment the Linear Contact Model with a Residual MLP}
\begin{figure}[h]
\vspace{-0.1in}
    \centering
    \includegraphics[height=0.35\linewidth]{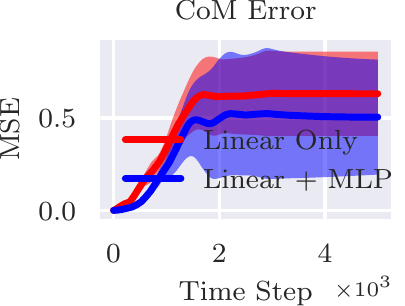}
        \includegraphics[height=0.35\linewidth]{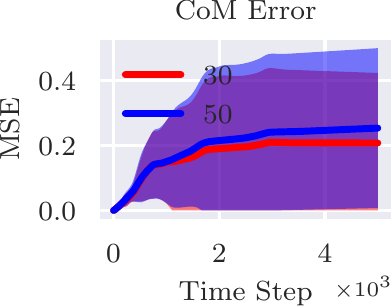}
    % \vspace{-8mm}
    \caption{Augmenting our linear contact model with a MLP could result a slightly better model if trained with 5 trajectories (left), but the performance doesn't consistently improve with 30 and 50 trajectories.}
    \label{fig:superball_linear+mlp_contact_model}
    \vspace{-2mm}
\end{figure}

The next idea is to accompany the proposed linear contact model with an MLP responsible to learn residual error terms. This can help learn complex dynamics as a complement to the drawbacks of the linear model. Figure ~\ref{fig:superball_linear+mlp_contact_model} shows the resulting performance. 

After 15 epochs, the training error keeps decreasing down to 0.05, which is much smaller than the linear model training error, 0.13. The testing CoM error, 0.5m, is smaller as well, which is only 21\% in the order of the length average 2.35m. This version requires only 5 training trajectories for training. If the training trajectories are increased to 30 and 50, the error is reduced in earlier time steps but later on it increases again. This is because the neural network faces difficulty in modeling the static states, where \emph{exact} normal force prediction is needed as counterpart of gravity. The MLP's prediction deviates from it, which leads to a drifting robot. Possible reasons are 1) the inputs of MLP have different magnitudes, but traditional methods to avoid over-fitting, e.g. normalization, violate the physical property. 2) MLPs are still too simple to learn contact responses and neural networks tailored to this task are still needed. 3) 30 and 50 trajectories are still not sufficient but more trajectories may also increase the chances of overfitting as well as increase human effort in a sim2real transfer. 

% \textcolor{red}{Adding more trajectories would continue to improve the performance after augmenting the linear contact model with a residual MLP.} This indicates that the model can be learned in a lifelong mode. The linear model also helps to get rid of overfitting of MLP.

% Since the linear model has a different learning rate to the MLP, we still apply a progressive training method here. We first learn our linear contact model and fix it. Then, we learn a MLP with a smaller learning rate. The residual MLP actually learns the 'error' of the linear model. Here the learning rate of MLP is 0.01 and is reduced every 4 epochs.

\subsection{SUPERball Wheel Rolling Open-loop Control}
This experiment executes iLQG (iterative Linear Quadratic Gaussian) both on MuJoCo and the differentiable engine. The evaluation is in terms of 1) data requirements: number of training trajectories and 2) policy transfer: difference in policies trained on the engine and on MuJoCo.

The task is to control the SUPERball rolling to the right at speed $10$. The setup for iLQG is as follows. We set the target linear velocity of CoM to $[10, 0, 0]$, the weight of each dimension is $[1, 0.5, 0.1]$, which means we prioritize the motion on x axis, allow swings on y axis and hope to maintain its height on z axis. The weight of KL divergence is 0.005. We use the exactly same setup for MuJoCo and our differentiable simulator. The linear contact model is trained with only one trajectory in this section. MLP is not included because it cannot predict static states accurately and the robot drifts at the initial static state.

\subsubsection{Data Requirement} The iLQG is trained in an iterative way. We run 20 iterations in total, sample 40 trajectories in each iteration and each trajectory is 5s, i.e. 5,000 time steps. So if we directly train the policy on the real platform, it needs $20\times40\times5,000=4M$ time steps. But our engine only needs a 5s trajectory to identify non-contact parameters and a 5s trajectories to identify contact parameters, i.e. $(1+1)*5,000=10k$ time steps, which is only \textbf{0.25\%} of $4M$.

\begin{figure}[h]
\vspace{-0.1in}
    \centering
    \includegraphics[height=0.2\linewidth]{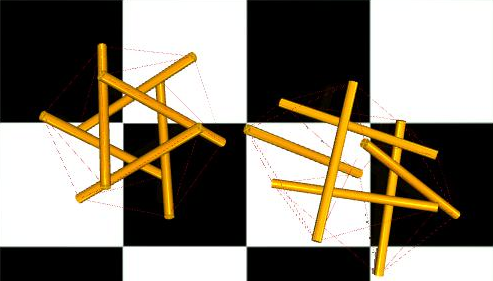}
    \includegraphics[height=0.2\linewidth]{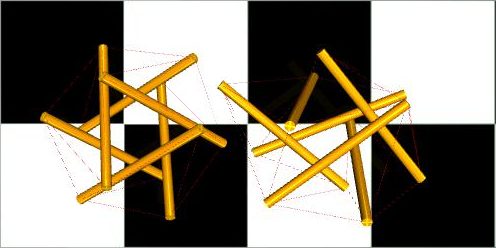}
    % \vspace{-8mm}
    \caption{The policy generated from our engine (right) for SUPERball is very close to the ground truth policy trained directly on MuJoCo (left).}
    \label{fig:policy_transfer_trajectory_comparison}
    \vspace{-2mm}
\end{figure}

\subsubsection{Policy Transfer}
\begin{figure}[h]
\vspace{-1mm}
    \centering
    \includegraphics[height=0.33\linewidth]{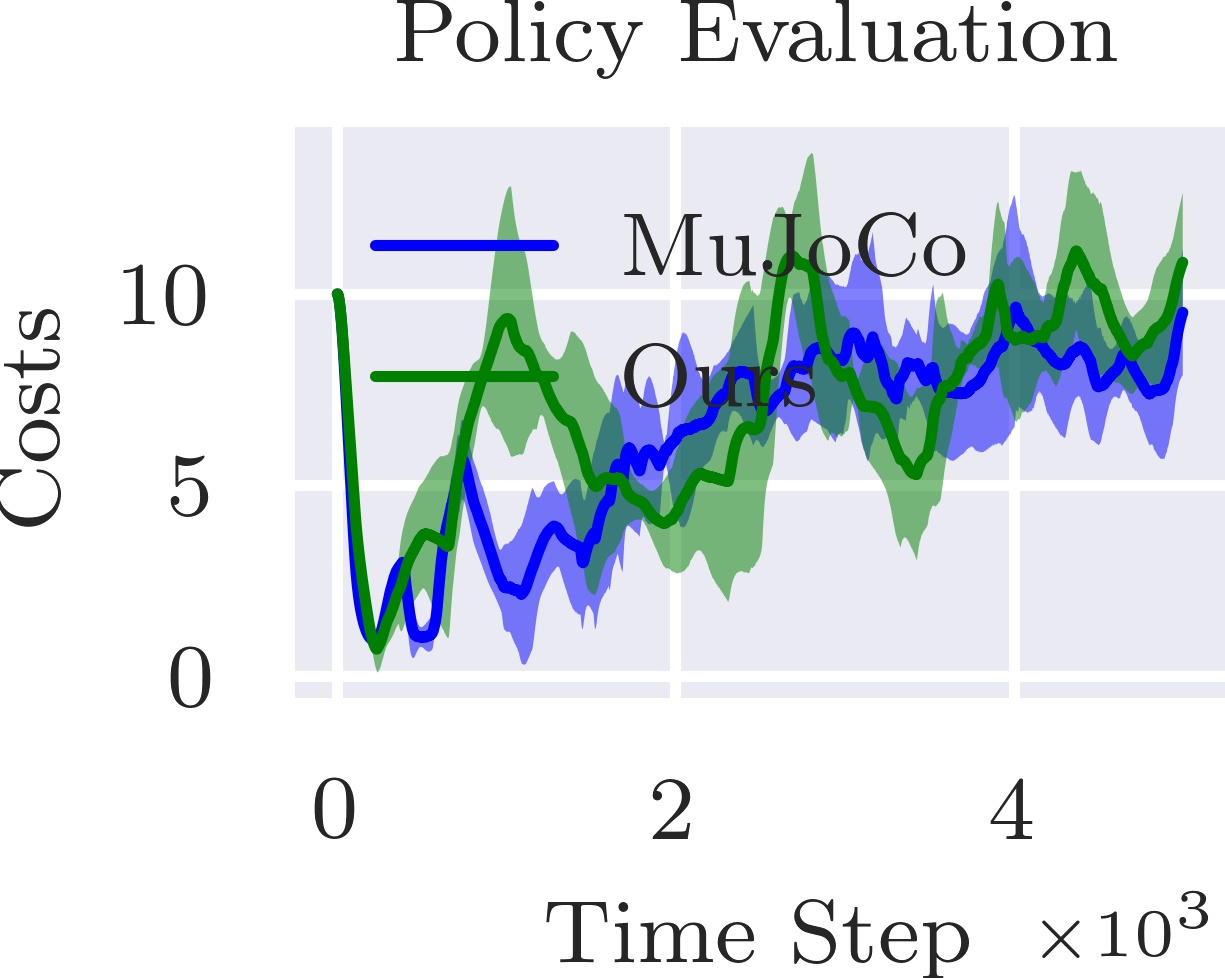}
    \includegraphics[height=0.33\linewidth]{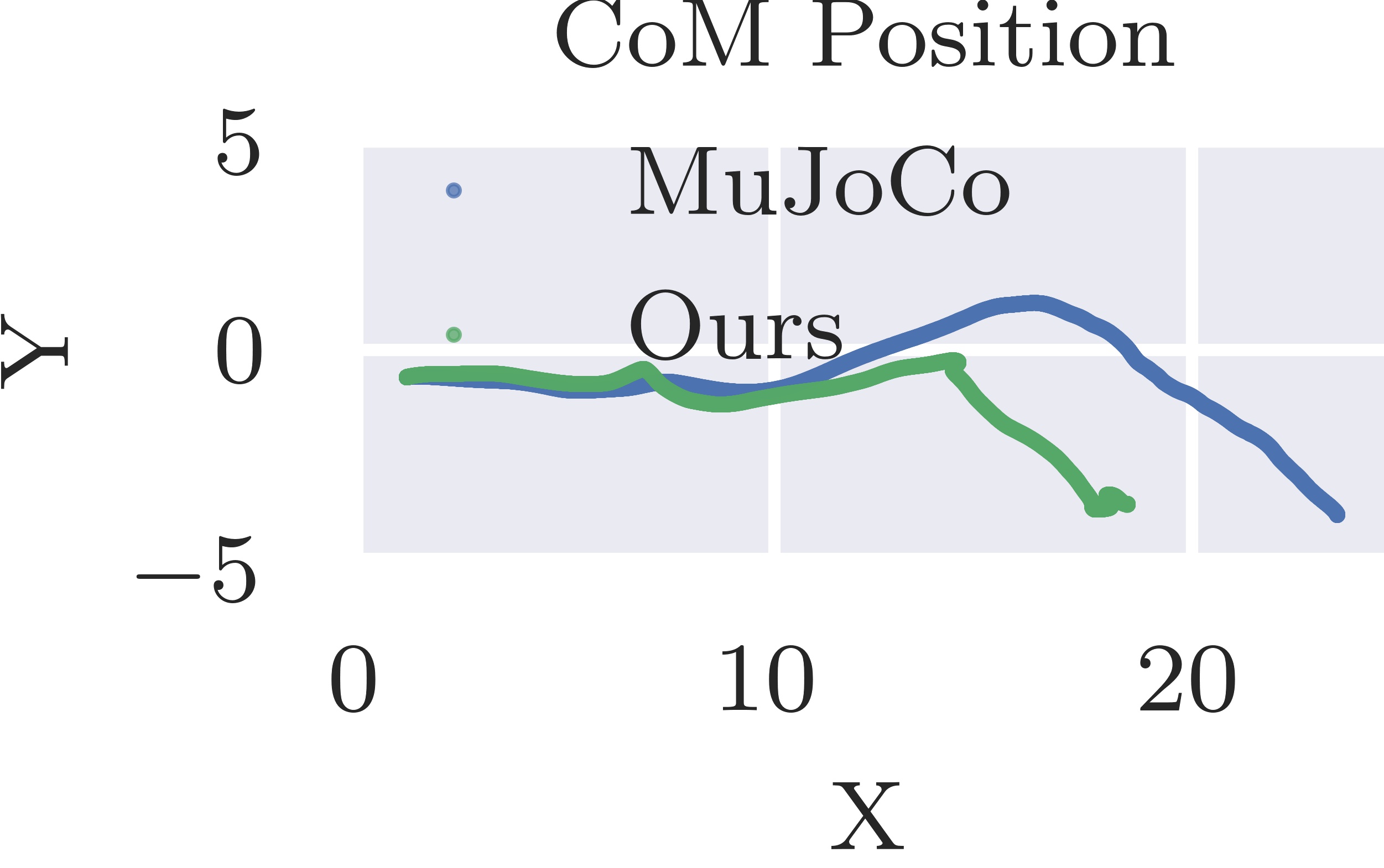}
    \vspace{-3mm}
    \caption{Comparison of policy for SUPERball learned on the diff. engine vs. that trained on the ground truth system: (left) trajectory cost on each time step; (right) projection of CoM onto the XY plane.}
    \label{fig:policy_transfer_cost_comparison}
    \vspace{-5mm}
\end{figure}

We run iLQG on MuJoCo directly to obtain the ground truth policy. Then we run iLQG on our engine to get our policy. We run iLQG 5 times for 5 policies on each and compute the mean and std. Finally, we evaluate the ground truth policy and the learned policy on the MuJoCo platform and compare their performance. We sampled a trajectory from each policy and computed the difference to the target linear velocity. Figure~\ref{fig:policy_transfer_trajectory_comparison} shows the trajectories from the two policies. The trajectory costs in Figure~\ref{fig:policy_transfer_cost_comparison} (left) and the mean trajectory projection onto the ground XY plane (right) indicate that these policies are very close. We don't apply any online adaptation for policy transfer. Thus our identified simulation is already good enough to generate good policies.

\vspace{-1mm}

\subsection{3-Bar Tensegrity Crawling Open-loop Control}

To evaluate the generality of the engine, we apply the learned engine to a different robot and new environment without adaptation. We introduced a 3-bar tensegrity robot, which is composed of 3 rods and 9 cables. We can only control 3 out of the 9 cables, which makes the problem harder. We also introduced a smaller gravity environment by reducing the gravity coefficient to $0.1g$, which is close to that on smaller planets. 

\begin{figure}[h]
\vspace{-0.1in}
    \centering
    \includegraphics[height=0.23\linewidth]{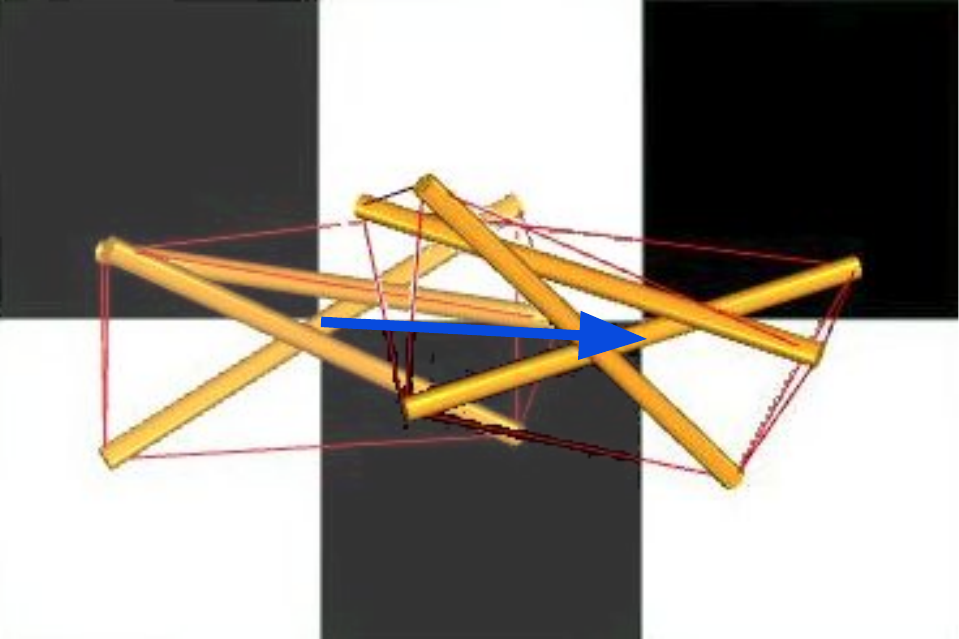}
    \includegraphics[height=0.23\linewidth]{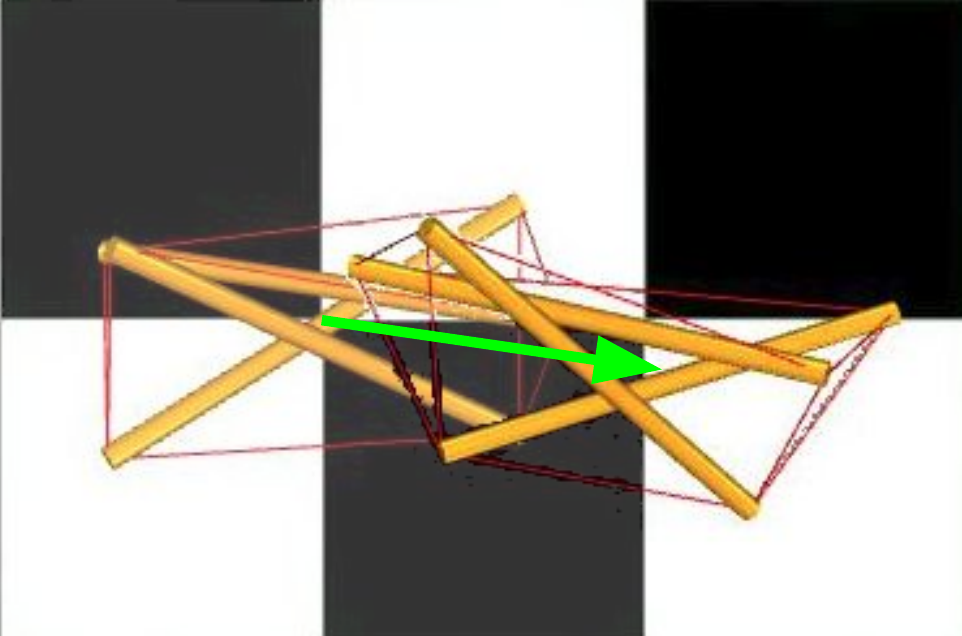}
    % \vspace{-8mm}
    \caption{The policy generated from the engine (right) for a 3-bar tensegrity  is very close to the ground truth policy directly trained on MuJoCo (left).}
    \label{fig:3bar_policy_transfer_trajectory_comparison}
    \vspace{-4mm}
\end{figure}

After running iLQG for 20 iterations, the engine can still generate competitive policies. It is because the new robot shares the same physical properties to the SUPERball that has different topology, which means we can still apply our learned physical parameters without any adaptation. The engine can also accept any changes in environment constants, e.g. gravity, which shows another advantage of explainable feature for policy transfer.

\begin{figure}[h]
\vspace{-.1in}
    \centering
    \includegraphics[height=0.4\linewidth]{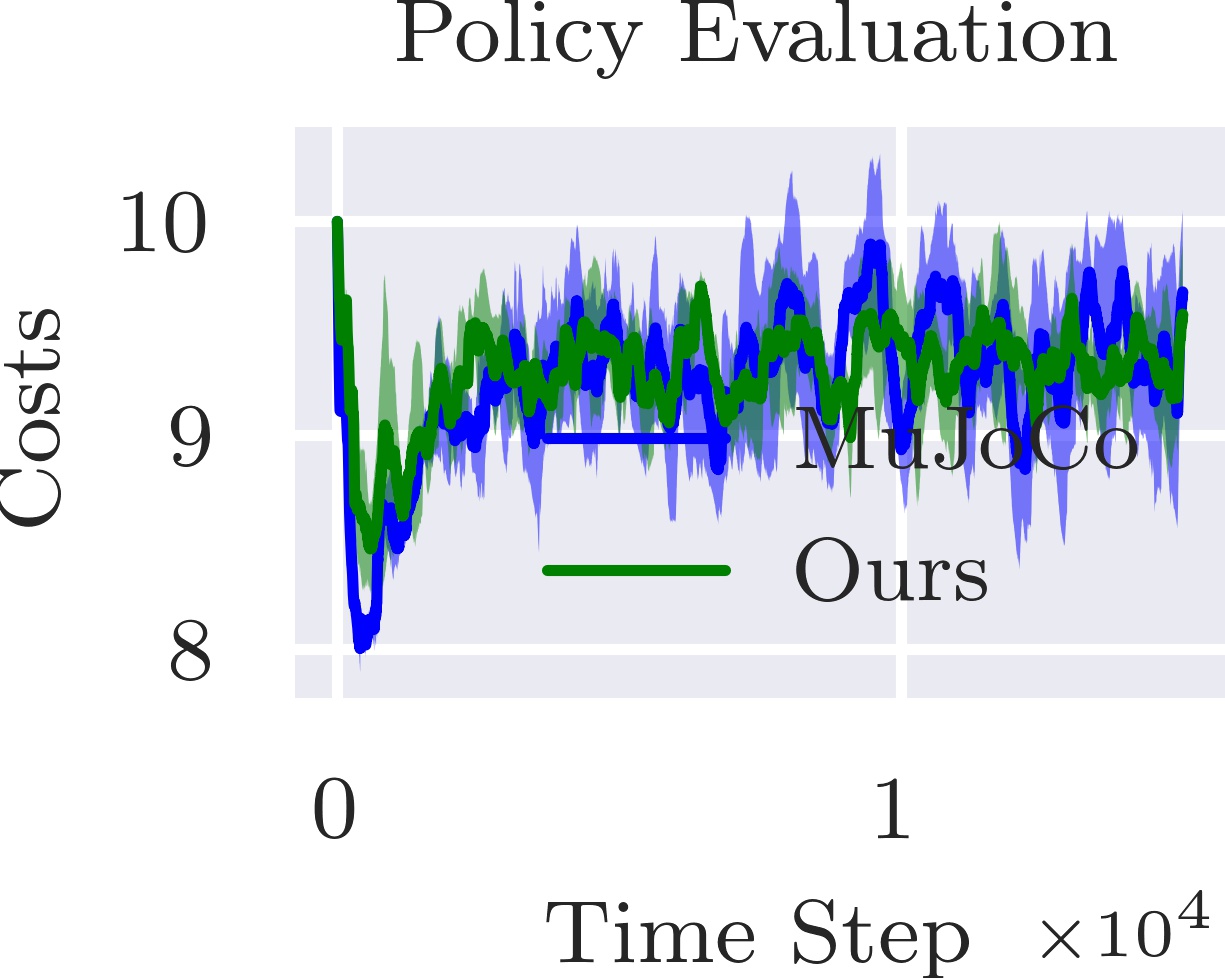}
    \includegraphics[height=0.4\linewidth]{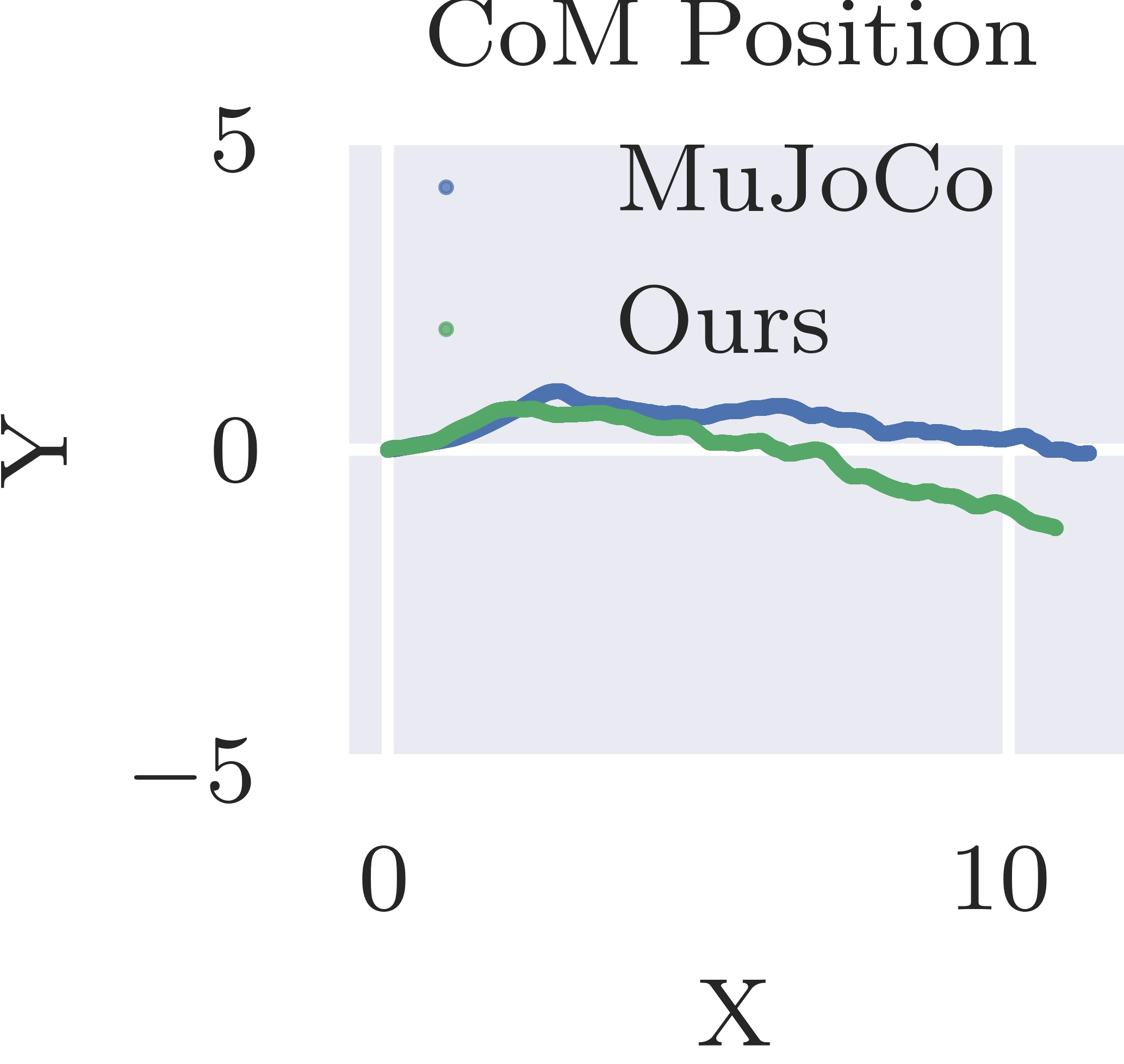}
    \vspace{-3mm}
    \caption{Comparison of policy for 3-bar tensegrity learned on the diff. engine vs. that trained on the ground truth system: (left) trajectory cost on each time step; (right) projection of CoM onto the XY plane in meter.}
    \label{fig:3bar_policy_transfer_cost_comparison}
    \vspace{-4mm}
\end{figure}

The engine is able to generate longer horizon controls. Here each trajectory component is 15s long. The control signal is sent every 100ms and 3 cables are manipulated, which includes $15,000\times3/100=450$ control signals. Generating the ground truth policy needs $20\times40\times15,000=12M$ data points. Even consider the 2 ground truth trajectories sampled for the SUPERball bot, it only needs $10k$ data points, which is only 0.083\% of $12M$. Thus, the long control policy benefits even more from the engine.

\section{CONCLUSION}
This paper proposed an end-to-end differentiable simulator for tensegrity robots. This simulator is composed only of simple linear models and all parameters are explainable. This helps to reduce its data requirement for system identification, which is critical to mitigate the reality gap in practice. The explainable parameters are easy to understand and their features are amenable to gradient descent. We also proposed a progressive training process to get rid of local optima and achieve higher accuracy for more confident parameter evaluation. The identified engine has been tested by generating policies, which are effective when applied on the original platform without adaptation.

%%%%%%%%%%%%%%%%%%%%%%%%%%%%%%%%%%%%%%%%%%%%%%%%%%%%%%%%%%%%%%%%%%%%%%%%%%%%%%%%

\bibliographystyle{IEEEtrans}
\bibliography{references}

\end{document}